\documentclass[11pt,a4paper]{article}
\usepackage[hyperref]{acl2017}
\usepackage{times}
\usepackage{latexsym}

\usepackage{url}

\usepackage[utf8]{inputenc}
\usepackage[bulgarian,english]{babel}

\usepackage{graphicx}
\usepackage{todonotes}
\usepackage{booktabs}

\aclfinalcopy 

\title{A Morpho-Syntactically Informed LSTM-CRF Model\\ for Named Entity Recognition}

\author{Lilia Simeonova \\
  FMI \\
  Sofia University \\
  Bulgaria \\
  {\tt lilia.valentinova.}\\ {\tt simeonova}\\ {\tt @gmail.com} \\\And
  Kiril Simov,  Petya Osenova \\
  LMaKP \\
  IICT-BAS \\
  Bulgaria \\
  {\tt \{kivs,petya\}}\\ {\tt @bultreebank.org} \\\And
  Preslav Nakov \\
  Qatar Computing  \\
  Research Institute, HBKU \\
  Qatar \\
  {\tt pnakov@qf.org.qa} \\
}

\date{}

\begin{document}
\maketitle
\begin{abstract}
We propose a morphologically informed model for named entity recognition, which is based on LSTM-CRF architecture and combines word embeddings, Bi-LSTM character embeddings, part-of-speech (POS) tags, and morphological information. 
While previous work has focused on learning from raw word input, using word and character embeddings only, we show that for morphologically rich languages, such as Bulgarian, access to POS information contributes more to the performance gains than the detailed morphological information.
Thus, we show that named entity recognition needs only coarse-grained POS tags, but at the same time it can benefit from simultaneously using some POS information of different granularity.
Our evaluation results over a standard dataset show sizeable improvements over the state-of-the-art for Bulgarian NER.
\end{abstract}

\section{Introduction}

\label{sec:introduction}

Although in recent years the {\em Named Entity Linking} (also known as Named Entity Disambiguation) task has been central in NLP research, the \emph{Named-entity recognition} (NER) task has remained far from solve, having in mind the productivity of names and the amount of information available in the era of big and noisy data. 

NER plays a critical role in the processing of texts with application to many real-world Natural Language Processing (NLP) tasks such as Question Answering, Information Extraction, Machine Translation, Dialog Systems, and chatbots, where it is sometimes called \emph{Concept Segmentation and Labeling} \cite{saleh-EtAl:2014:Coling}.

\noindent Traditionally, NER has focused on recognizing entities such as person (PER), organization (ORG), location (LOC), and miscellaneous (MISC). This tradition goes back to the Message Understanding Conference (MUC) for English~\cite{Grishman:1996:MUC:992628.992709}, and the subsequent CoNLL 2002/2003 shared tasks, which also targeted other European Languages such as Spanish, Dutch, and German 
\cite{TjongKimSang:2003:ICS:1119176.1119195}.\footnote{Other schemata such as ACE~\cite{doddington-etal-2004-automatic} used a richer inventory of entity types.} This same setup was followed in more recent work for a number of other languages, and we also follow it in the present work.

Early systems relied on hand-crafted rules with pattern-matching \cite{Appelt:1995:SIF:1072399.1072420}. Unfortunately, this required an large pre-annotated datasets, collecting which was time-consuming and error-prone. The next step was to add gazetteers and lexicons that were generated automatically or semi-automatically \cite{Popescu:2005:EPF:1220575.1220618}. Adding such resources required special approaches to resolve the ambiguity between names and common words. Such problems were solved using models such as Hidden Markov Models \cite{Zhou:2002:NER:1073083.1073163} and Conditional Random Fields \cite{Sutton:2012:ICR:2688180.2688181}.

In our work here, we use deep neural networks for Bulgarian NER. \citet{N16-1030} have shown remarkable results for English, using a combination of Bi-LSTMs (Bi-directional Long Short-Term Memory) and CRF. However, the approach is problematic for morphologically rich languages. The main problem is the missing information within word embeddings for the numerous word forms involved in multiword names that require additional grammatical knowledge in order to be processed properly. Here we incorporate such information as additional input to our neural model.

Our contributions are as follows:
\begin{itemize}
	\item We show that for morphologically rich languages	such as Bulgarian the access to POS and morphological annotation is crucial and can yield very sizeable performance gains.
    \item We achieve sizable improvements over the state-of-the-art for Bulgarian NER.
    \item Finally, we make our data and code freely available, which should enable direct comparison in future work.\footnote{\url{http://github.com/lilia-simeonova/NER-bg/}}
\end{itemize}

\section{Related Work}
\label{sec:related}

Our work is based on Bulgarian, but we claim that it is appropriate also for other languages with rich morphological systems like Slavic and Romance languages, for example. For that reason, we present first the best results for NER in other Slavic languages having in mind that they are synthetic, while Bulgarian is a predominantly analytic language whose morphological richness lies exclusively in the verbal system and not so much in the nominal one. Analytism implies more types of multiword named entities in Bulgarian but less inflection variety, and different distribution of the common types for these languages. The direct comparison of the numbers presented below should be taken with a grain of salt as they are on different datasets and for different languages. Yet, they are indicative for the different methods used for these languages.

For Russian, a Hybrid Bi-LSTM approach was applied by \citet{10.1007/978-3-319-71746-3_8}, who achieved precision of 89.57, recall of 84.89, and F1 score of 87.17. These results are comparable to the ones by our model using the same approach.

For Czech, \citet{10.1007/978-3-642-40585-3_10} reported precision of 88.27, recall of 78.00, and F1 score of 82.82 using a Maximum Entropy Markov Model. The feature modeling also proved to be working in Czech, as their best results used features based on morphological analysis, two-stage prediction, word clustering, and gazetteers.

For Polish, \citet{10.1007/978-3-540-39985-8_24} achieved precision of 91.0, recall of 77.5, and F1 score of 82.4. They used the SProUT system, which is an NLP platform, consisting of pattern/action rules.

\noindent In the last years, the interest in NER for Slavic languages grew. Two shared tasks were organized ---- the first and the second Multilingual Named Entity Challenge in Slavic Languages. They have been descibed in \cite{DBLP:conf/acl-bsnlp/PiskorskiPSSY17} and \cite{piskorski-etal-2019-second}. The challenges included several tasks: recognition of mentions of named entities in Web documents in seven Slavic languages (Bulgarian, Croatian, Czech, Polish, Russian, Slovak, Slovene, Ukrainian), their normalization/lemmatization as well as cross-lingual linking. 

Our evaluation on NER in this paper is more similar to the relaxed evaluation parameter where the string is detected and classified, not the invariant. Considering the complexity of the task, the drop of the results per language and per entity types have been expected. Such a task, however, is also good motivation for improving the NER systems for Slavic languages, including Bulgarian.

There is some previous work on NER for Bulgarian. \citet{Georgiev2009FeatureRichNE} presented a model using Conditional Random Fields with several hand-crafted features. They combined well-established features used for other languages with language-specific lexical, syntactic, and morphological information. Their result is the previous state-of-the-art for Bulgarian.

So far, the highest reported results for NER are for English. For example, \citet{Chiu2016NamedER} reported an F1 score of 91.20 using Bi-LSTM + CNN + gazetteers + linking, while \citet{Passos2014LexiconIP} achieved an F1 score of 90.90 using a new form of learning word embeddings that can leverage information from relevant lexicons. For German, \citet{Gillick2016MultilingualLP} achieved an F1 score of 82.84, which shows that the rich morphology causes a drop in the performance.

Currently, the prevalent paradigm in NLP is to use neural networks, typically based on LSTMs or CNNs. 
As we have mentioned above, \citet{N16-1030} proposed an LSTM-CRF model for NER.\footnote{They also proposed a transition-based model inspired by shift-reduce parsers, but the results were worse.} The model uses a bi-directional LSTM to encode the left and the right context of the current input word. Then it passes the concatenation of the two hidden vectors (one produced by the left LSTM and one by the right LSTM) to a CRF model. Its task is to ensure the global consistency of the NER tags. 

\noindent In this model, each input word is represented as a concatenation of its word embedding and the character-level embedding for the word produced by a character Bi-LSTM. The character embedding provides features for the suffix and the prefix of the word. Thus, the left-to-right character-based LSTM embedding models the word suffix, while the right-to-left one models the word prefix.
The word embeddings are trained in an unsupervised manner on external data, while the character-based LSTM embeddings are trained on the training data as part of the end-to-end training of the full LSTM-CRF model. This model does not need any explicit feature engineering nor does it need manual gazetteers; yet, it achieved state-of-the-art performance for four languages: English, German, Dutch, and Spanish.
Here we take this model as a basis, and we augment it to model part-of-speech (POS) and grammatical information, which turns out to be very important for a morphologically complex language such as Bulgarian.

\citet{Strubell:2017} extended the above model by substituting the LSTM with Iterated Dilated Convolutional Neural Networks, a variant of CNN, which permit fixed-depth convolutions to run in parallel across entire documents, thus making use of GPUs, which yields up to 20-fold speed up, while retaining performance comparable to that of the LSTM-CRF model. They further aggregated context from the entire input document, which they found to be helpful. 
In our preliminary monolingual experiments, this model performed very similarly, but slightly worse, than the LSTM-CRF model, and thus we chose LSTM-CRF for our experiments below.

\section{Data}
\label{sec:data}

In this paper, we work with a Bulgarian corpus, annotated with BIO tags and positional tags, the same as in the CoNLL-2002 shared task \cite{TjongKimSang:2002:ICS:1118853.1118877}. 
The data is in BIO format, which encodes for each token in the text whether it is at the beginning of the expression of interest (Named Entity in our case), inside or outside of it. The annotation in the available Bulgarian data comes from the manually annotated Bulgarian treebank, known as BulTreeBank \cite{BulTreeBank}. In the treebank, each NE is represented as a constituent consisting of one or more tokens. Each NE phrase is annotated by the categories Person, Organization, Location, and Other. 

\begin{table}[tbh]
\small
  \centering
  \begin{tabular}{lllll}
  \toprule
   \begin{otherlanguage*}{bulgarian}Христо\end{otherlanguage*}  & \begin{otherlanguage*}{bulgarian}Стоичков\end{otherlanguage*}& \begin{otherlanguage*}{bulgarian}пристигна\end{otherlanguage*} & \begin{otherlanguage*}{bulgarian}в\end{otherlanguage*}& \begin{otherlanguage*}{bulgarian}София\end{otherlanguage*}\\
Hristo & Stoichkov & arrived & in & Sofia \\
\midrule
B-PER & I-PER & O & O & B-LOC  \\
\bottomrule
  \end{tabular}
  \caption{\label{bio-encoding}An example in BIO encoding.}
\end{table}

The BIO tags for the tokens forming a given named entity (NE) in the treebank are created on the basis of the syntactic annotation. The first word in the phrase is marked as the beginning of the NE, while the rest of the tokens ar emarked as inside of the NE. The tokens that are not part of any NE are encoded as outside elements. 
In order to represent the category of the NE, each tag for begin and inside tokens includes a modifier for the category. Thus, we use nine labels: B-PER, I-PER, B-ORG, I-ORG, B-LOC, I-LOC, B-MISC, I-MISC, O. The example in Table~\ref{bio-encoding} shows a simple sentence annotated with two named entities: a person name (\emph{Hristo Stoichkov}) and a location name (\emph{Sofia}). 

Besides the BIO tagging, the texts in the dataset inherited the morphosyntactic annotation from the treebank. This annotation uses the BulTreeBank Morphosyntactic Tagset \cite{BTB-TR03}. The tagset is positional. It encodes parts-of-speech and grammatical features for Bulgarian. For example, {\em Npfsi} stands for noun, proper, feminine, singular, indefinite. This annotation offers an opportunity to explore how the morphological features can affect NER. 

The resulting dataset is divided into three disjoint sets: training set (Train), development set (Dev), and test set (Test). Table~\ref{named-entity-tokens} shows statistics about the annotated data.
We can see that a large number of examples are labeled as Person names, and that the distribution of Locations, Persons and Organizations is not balanced. While we can still build a stable system based on this data, the class imbalance makes our model more vulnerable to overfitting. Thus, we use early stopping in order to prevent the model from continuing to learn weights and parameters if it does not see an improvement in the final score.

\begin{table}[tbh]
  \small
  \centering
  \begin{tabular}{@{ }@{ }l@{ }@{ }r@{ }@{ }r@{ }@{ }r@{ }@{ }r@{ }@{ }r@{ }@{ }r@{ }@{ }}
           \toprule
         & \bf Sent. & \bf Tokens & \bf PER & \bf ORG & \bf LOC & \bf MISC     \\
    \midrule
    Train & 28,636& 528,567  & 16, 804 & 3,028 & 6,786 &  911   \\
    Dev  & 4,063 & 64,014 & 2,514 & 515 & 1,021  &  227 \\
    Test & 3,907 & 60,645  & 1,875 &  305 & 781 & 112  \\
	\bottomrule
  \end{tabular}
  \caption{\label{named-entity-tokens}Statistics about the data.}
\end{table}

\section{Model}
\label{sec:method}

As mentioned above, we construct our model as a modification of the Bi-LSTM-CRF architecture from \cite{N16-1030}. After some experiments with the original system, we decided to modify its input: we added a vector representing some of the information encoded in the morphosyntactic tags. Thus, we created the input vectors for the tokens in the sentences as a concatenation of three vectors: a word embedding vector, a character embedding vector, and a vector containing some grammatical features, called a {\em grammatical vector}. We experimented with different grammatical vectors, as explained below.

The rest of the Bi-LSTM-CRF architecture of \cite{N16-1030} was kept as in the original model: First, we run a Bi-LSTM over the sequence of word vectors so that we could get their contextual word representation. We use a fully connected neural network to get a score for each of the tags. At the end, we run a CRF decoder to decide what the best combination of scores is. 

The key takeout of our model is that we use some feature modeling to show that for a morphologically rich language such as Bulgarian using POS and grammatical information can improve the results. Thus, we mix automatically learned features --- the word and the character embeddings ---, with hand-crafted features encoded as a grammatical vector.  

In the rest of this section, we describe the different components of our system.

\paragraph{LSTM-CRF Implementation}
For the implementation of the general LSTM-CRF architecture, we use Tensorflow \cite{DBLP:journals/corr/SakSB14}.

\paragraph{Word Embedding}
Nowadays there are many different approaches to train word vectors such as Word2vec \cite{DBLP:journals/corr/abs-1301-3781}, GloVe \cite{pennington2014glove}, FastText \cite{BojanowskiGJM16}, and many more. In our experiments, we use the pre-trained Bulgarian word embeddings from FastText \cite{BojanowskiGJM16}.\footnote{In this work, we do not use any contextualization of the word embeddings such as ElMo \cite{Peters:2018} and BERT \cite{devlin2018bert}, as our Bi-LSTM architecture already performs contextualization.} This choice was motivated by the fact that FastText uses the structure of the words by taking into consideration character $n$-grams, thus modeling morphology and many out-of-vocabulary words.

\paragraph{Character Bi-LSTM Embedding}

In order to produce character embeddings, we use a bi-directional LSTM over the character representation of the text. For each character in the text, each of the two LSTMs produces an hidden vector. For each word, the hidden vector for the last character produced by the left-to-right LSTM models information about the suffix of the word. Similarly, the hidden vector for the first character produced by the right-to-left LSTM models information about the prefix of the word. Following the approach, used in \cite{DBLP:conf/emnlp/LingDBTFAML15}, we constructed the final character embedding of the word as a concatenation of the prefix and the suffix vectors.

\paragraph{Grammatical Vectors}
We use several types of grammatical vectors or their combinations. They are divided into POS vectors that encode different combinations of parts-of-speech and morphological vectors encoding other grammatical features.

\paragraph{\em POS Vectors}

The part-of-speech information for each word is represented as an a one-hot vector with eleven positions. This vector is concatenated to the vector for the word embedding. In the tagset, we have the following parts-of-speech: {\bf N} --- noun, {\bf A} --- adjective, {\bf V} --- verb, {\bf H} --- hybrid, {\bf D} --- adverb, {\bf R} --- preposition, {\bf P} --- pronoun, {\bf C} --- conjunction, {\bf T} --- particle, {\bf M} --- numeral, and {\bf I} --- interjection. In our experiments, we divided these parts-of-speech into different groups depending on the role they play in the representation of the named entities. For example, the tags A, N, H, R were viewed as a possible part of a named entity in contrast to the others that cannot form named entities. In this case, the one-hot vector contains only two positions. The  groups are given in the experimental section below.
The Hybrid tag (H) is special in the tagset. It refers to both family names and name adjectives. Bulgarian family names (as other Slavic ones) are proper names, but morphologically they behave like adjectives due to their adjectival origin. 

\paragraph{\em Morphological Vectors}

The nominal system of Bulgarian shares some features with other Slavic languages, such as agreement in grammatical gender and number, rich pronoun system, etc. However, it has also specific features, such as the post-positioned definite article and lost nominal declension system. Our aim is to show the contribution of all these types of features to the named entity recognition task.

\begin{figure}[htb]
  \centering
  \includegraphics{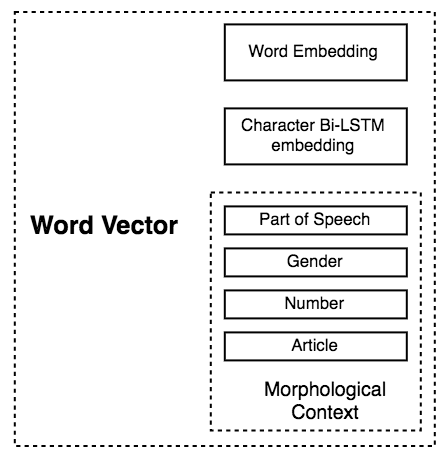}
  \caption{Full input vector representation with concatenation of word embeddings, character embeddings, and morphological features.}
  \label{fig:example}
\end{figure}

Another thing worth mentioning is that the nouns, adjectives and hybrid tags share some common features. This information appears to be very useful for recognizing the more specific types of named entities. Note that the existence of a preposition in a sequence can distinguish some further patterns as well.

The morphological features vary between the different entities, but there are few that can be defined for nouns, adjectives, hybrid tags and pronouns. Some of them are arguably useful, such as gender, number, and definiteness, and we will describe them briefly here:

\begin{description}
\item[Gender] can have three values: \emph{masculine, feminine, and neutral};
\item[Number] can have four values: \emph{singular, plural, only plural} and \emph{count form} (which is only for masculine nouns for non-persons)'
\item[Definiteness] can have four values: \emph{indefinite, definite, short definite} and \emph{full definite}. The latter two are for singular masculine nouns only.
\end{description}

For each word with a POS tag of noun, pronoun, adjective or hybrid, we concatenate a one-hot vector representation for each of the features above (we use a zero vector for the rest).
We form the final version of our word vector as a concatenation of all contextual vectors, as shown on Figure~\ref{fig:example}.

\paragraph{Dropout}
In order to prevent overfitting, we use a dropout layer on top of our word vectors as described in \cite{DBLP:journals/corr/abs-1207-0580}. For each presentation of each training example, we randomly exclude a hidden unit from the network with a certain probability. In this way, the system learns to detect and use more useful features.

\begin{table*}[tbh]
  \small
  \centering
  \begin{tabular}{lcccc}
\\
\toprule
   & \bf No Morphology & \multicolumn{3}{c}{\bf With Morpholofy}\\
   \bf Model  & \bf F1  & \bf F1 & \bf P  & \bf R \\
\midrule
    Model + POS-2  & 90.04 & 90.44 &     92.02 &  88.90    \\
     Model + POS-3  & 91.20 & 91.11 & 91.66 &  90.57   \\
    Model + POS-4  & 90.16& 90.83 & 92.30 & 89.41   \\ 
    Model + POS-5 & 91.32 & 90.58   & 92.42 &  88.82   \\  
    
    Model + POS-11 &90.96 &91.03& 91.60 & 90.48     \\

 \midrule
 Model + POS-3+11 & 91.18 & 92.20 & 93.31&  91.12 \\ 

    Model + POS-4+11 & 90.89 & 91.04  &92.17 & 89.94    \\
	\bottomrule
  \end{tabular}
  \caption{\label{ablation-pos}Evaluation results for Bulgarian POS tagging. Shown are results where the standard input to the Bi-LSTM-CRF model is augmented with different POS tags and morphological features.}
\end{table*}

\begin{table*}[t!]
  \small
  \centering
  \begin{tabular}{llllllll}
  \\
\toprule
  \multicolumn{7}{l} {English}{Ivan Valtchev visited the Bulgarian Academy of Sciences.}\\

  \multicolumn{7}{l} {Bulgarian}{\begin{otherlanguage*}{bulgarian}Иван Вълчев посети Българската академия на науките.
\end{otherlanguage*}} \\
\midrule
   Tokens   & \begin{otherlanguage*}{bulgarian}Иван\end{otherlanguage*}  & \begin{otherlanguage*}{bulgarian}Вълчев\end{otherlanguage*}& \begin{otherlanguage*}{bulgarian}посети\end{otherlanguage*} & \begin{otherlanguage*}{bulgarian}Българската\end{otherlanguage*} & \begin{otherlanguage*}{bulgarian}академия\end{otherlanguage*} & \begin{otherlanguage*}{bulgarian}на\end{otherlanguage*}& \begin{otherlanguage*}{bulgarian}науките\end{otherlanguage*}\\
\midrule
POS11 & N & H & V & A &  N & R & N \\
POS2 & ANHR & ANHR & O & ANHR & ANHR & ANHR & ANHR \\
POS3 & ANH & ANH & O & ANH & ANH & R & ANH\\
POS4 & NH & NH & O & A & NH & R & NH\\
POS5 & N & H & O & A & N & R & N \\
\bottomrule
  \end{tabular}
  \caption{\label{prepoc-schemes}Example of the POS tags for annotation schemes of different granularities when applied to the same Bulgarian sentence.}
\end{table*}

\section{Experiments and Evaluation}
\label{sec:experiments-eval}

In this section, we present the experimental setup and the evaluation results for the different models we experimented with.

\subsection{Training and Hyper-parameters}

We experimented with different values of the hyper-parameters and we found that changing some of them can result in sizeable improvements. The most considerable difference was for the learning method and the learning rate. Our best resulst were achieved using the Adam optimizer \cite{DBLP:journals/corr/KingmaB14}, which is computationally efficient and has minimal memory requirements. 

At the beginning, we set the learning rate to the initial value of 0.001, and then at each epoch, we multiplied it by a specific learning decay value. Decreasing our learning rate over time can help us find the minimum of our function without actually missing it. While Adam already decays the learning rate at each iteration, previous work has found that tuning the initial learning rate could yield sizeable improvements over the default settings. Thus, we use this additional decay. \cite{DBLP:conf/nips/WilsonRSSR17}

The word embeddings we use from FastText have a dimensionality of 300, while the character embedding vectors have a dimensionality of 100. In order to produce them, we uses the TensorFlow default Xavier initializer and then we ran a Bi-LSTM on top of them in order to obtain contextual vectors with no additional layers. 

We set the batch size  to 20 and the dropout to 2. The Adam's parameters we used are as follows: $learning rate=0.001$; $\beta1=0.9$; $\beta2=0.999$;
$\epsilon=1e-08$; 
$use\_locking=False$. We also added a gradient clipping with a value of 1.

At decoding time, we used a linear-chain CRF \cite{Lafferty01conditionalrandom}. This model has been shown to outperform a simple SoftMax classifier as the tagging decision needs to be global.

\subsection{Experiments}

The experimental results suggest that adding grammatical features can have a sizeable impact on the performance of the general LSTM-CRF model for Bulgarian NER. In Table \ref{ablation-pos}, we can see an example of different combinations of POS tags, where \emph{Model} stands for Bi-LSTM-CRF and POS represents a one-hot encoding for the following:

\begin{description}
\label{sec:pos-schemes}
\item[POS11] = all part of speech tags separately
\item[POS2] "ANHR" vs. REST
\item[POS3]"ANH" vs. "R" vs. REST
\item[POS4] "A" vs. "NH" vs. "R" vs. REST
\item[POS5] "A" vs. "N" vs. "H" vs. "R" vs. REST

\item[POS3 + POS11] POS11 vs. "ANHR" vs. REST
\item[POS4 + POS11] POS11 vs. "ANH" vs. REST
\item[Morph] Gender, number, and definiteness
\end{description}

We further perform several experiments in order to determine whether we need the full set of part-of-speech tags or it is enough just to know whether the entity is part of the group of the nouns, adjectives, hybrid tags, and prepositions. 

In Table \ref{prepoc-schemes}, we see how the entities map to the different POS groups.
It appears that knowing the concrete tag of each entity can help us improve the performance by almost one percent.

Table \ref{model-comparisons} shows our best result compared to the previous state-of-the-art result as reported in \cite{Georgiev2009FeatureRichNE}. They achieved an F1 score of 89.4\% by sing regular expressions, gazetteers and non-local morpho-syntactic characteristics. Our model improves this to 92.20\% without using any external resources.

\begin{table}[t!]
  \small
  \centering
  \begin{tabular}{ll}
\\
\toprule
   \bf Model     & \bf F1 \\
   \midrule

    \cite{Georgiev2009FeatureRichNE} & 89.40      \\
    Our model     & 92.20  \\
\bottomrule
  \end{tabular}
  \caption{\label{model-comparisons}Comparing the previous state-of-the-art results to our best morphologically informed Bi-LSTM-CRF model.}
\end{table}

More detailed evaluation results for our best model are presented in Table \ref{detailed-results}, where we show the precision, recall and F1 score for each type of named entity. We can observe relatively worse F1 score of 84.70 for Organization compared to 95.86 for Location and 94.95 for Person. We explain this drop in F1 score by the fact that many organizations are named after persons. 

\begin{table}[tbh]
  \small
  \centering
  \begin{tabular}{llll}
\\
\toprule
   \bf Entity     & \bf Precision &\bf Recall & \bf F1 \\
	\midrule
    Location & 97.75 & 94.05 & 95.86      \\
    Person  & 95.67 & 94.23 & 94.95  \\
    Organization  &75.57 & 96.34 &  84.70\\
    Miscellaneous & 96.15 & 22.73 & 36.76  \\

\midrule
    Overall  & 93.31 &  91.12 &  92.20 \\
\bottomrule
  \end{tabular}
  \caption{\label{detailed-results}Detailed results for the different kinds of named entities.}
\end{table}

\begin{table}[tbh]
  \small
  \centering
  \begin{tabular}{lc}
	\\
\toprule
\bf Model     & \bf F1 \\
	\midrule
    (1) LSTM-CRF (words only) & 82.03     \\
    (2) fwd-LSTM-char + (1) &  85.15      \\
    (3) bwd-LSTM-char + (1) &  85.40  \\
    (4) Bi-LSTM-char + (1) &  86.44  \\
    (5) POS11 + (4)   &  90.96  \\
    (6) Morph + (5) &  91.03  \\   
    (7) POS3 + (6)  &  92.20  \\
\bottomrule
  \end{tabular}
  \caption{\label{impact}The impact of different components and different component combinations on the performance of our best model.}  
\end{table}

Table~\ref{impact} shows the cumulative effect of adding different components to our model. The basic model we started with is shown on line (4). Then, on lines (1)-(3) we remove different components from this basic model, and on lines (5)-(7) we add POS and morphological information to it. We can see sizeable improvement for the standard Bi-LSTM-CRF model with only word vector representationa and the model with character-level LSTM. Interestingly, there is almost no difference between the suffix and the prefix vectors.
We can further see that adding POS11 (5) vector improves the performance by almost four percent absolute. The morphological vectors and POS3 also improved the F1 score to 92.20 points absolute. These improvements show that using known linguistic knowledge such as grammatical features could improve the representation vectors learned over huge text corpora. From the point of view of feature learning, we speculate that vectors trained over texts in morphologically rich languages do not learn enough grammar such as POS and morphology. The character embeddings also seem not to help much. One explanation for this could be that the suffixes and prefixes in Bulgarian are also highly ambiguous.

\begin{table}[tbh]
  \small
  \centering
  \begin{tabular}{ll}
	\\
    \toprule
   \bf Model     & \bf Word (Bulgarian / English) \\
\midrule
   	Model &\begin{otherlanguage*}{bulgarian}Еминем\end{otherlanguage*} / Eminem    \\
    Model + POS11     &\begin{otherlanguage*}{bulgarian}Фердинанд\end{otherlanguage*} / Ferdinand   \\
   	Model + POS2 & \begin{otherlanguage*}{bulgarian}Ваксберг\end{otherlanguage*} / Vaksberg \\
	Model + POS3 & \begin{otherlanguage*}{bulgarian}Гьоте \end{otherlanguage*} / Goethe \\
    Model + POS4 & \begin{otherlanguage*}{bulgarian}Обзървър \end{otherlanguage*} / Observer \\
   Model + POS5 & \begin{otherlanguage*}{bulgarian}Шехеразада \end{otherlanguage*} / Scheherazade \\
   \midrule
   Model + POS11 + POS3 & \begin{otherlanguage*}{bulgarian}Ингмар \end{otherlanguage*} / Ingmar \\
    \bottomrule
  \end{tabular}
  \caption{\label{error-analysis}Examples of words which could not be handled correctly by the specific configuration}  
\end{table}

\section{Error Analysis}
\label{sec:error-analysis}

Our manual analysis of the errors shows that one of the main reasons for our model to work better when POS tags are provided is due to the presence of many loanwords in the Bulgarian text. The LSTM-CRF model manages to learn the grammar of the language itself, but it needs additional help with words borrowed from other languages.

A common problem for the LSTM-CRF model is the mislabeling of foreign person or organization names. In such cases, the POS tags help by suggesting the possible part-of-speech for each word.
In our test set, around 10\% of the wrongly labeled words are loan words, borrowed primarily from English, Russian, German and Turkish. 
Table~\ref{error-analysis}, shows some examples of words that could not be handled properly.

\begin{table*}[tbh]
  \small
  \centering
  \begin{tabular}{rrrrrrrrrrrr}
\\
\toprule
      & B-ORG & I-ORG & B-PER & I-PER &  B-LOC & I-LOC & B-MISC & I-MISC & O \\
\midrule
B-ORG   & \bf 258 &    0   & 1   & 0       &  2        & 0     & 6 & 0 & 6 \\
I-ORG & 0 & \bf 31 & 0  & 0 &   0     & 0   & 0   & 0 & 1 \\
B-PER & 9 & 1 & \bf 1,169 & 6 & 8 & 0 & 7 & 0 & 58\\
I-PER & 0 & 1 & 15 & \bf 595 & 0 & 0 & 0 & 0 & 6\\
B-LOC & 15 & 0 & 8 & 0 & \bf 676 & 0 & 4 & 0 &36 \\
I-LOC & 1 & 0 & 0 & 1 & 3 & \bf 36 & 0 & 0 & 1\\
B-MISC & 27 & 0 & 2 & 0 & 1 & 0 & \bf 39 & 0 & 41\\
I-MISC & 0 & 1 & 0 & 0 & 0 & 0 & 0 & \bf 1 & 0\\
O & 12 & 2 & 10 & 22 & 2 & 0 & 3 & 0 & \bf 57, 522 \\

\bottomrule
  \end{tabular}
  \caption{\label{confusion-matrix}Confusion matrix for our best model on the test dataset: the columns represent the true labels and the rows show the predictions.}
\end{table*}

The loanwords cannot be successfully recognized by our algorithm in all cases. Even though some people try to write them in Cyrillic, their structure is different from the typical structure of the Bulgarian words. That is why, we further tried to add gazetteers and lexicons with existing loan words in Bulgarian. Similarly to the way we added POS vectors, we created a one-hot vector for each word that says whether that word is part of our lexicon with loan words or not. We then concatenated the vector to the rest of the word embeddings.  However, this approach did not result in significant improvements because new words are added to the language almost every day, and it is impossible to capture them all. 

Sometimes, the loanwords come in the Latin alphabet, as they are spelled in their original language. For such cases, we added a feature to the model that captures the information whether the words are in Latin or in Cyrillic.
This feature, by itself, did not make much of a difference. Yet, we plan to explore it further in our future work.

Table \ref{confusion-matrix} shows a confusion matrix for the nine BIO tags that we used for the four kinds of named entities that we are recognizing. In the table, the columns represent the actual expected gold tags, while the rows show the predictions of our model.
There are several interesting observations that we can make about this confusion matrix. First, it looks like the biggest problem for the model is with the tag B-PER, which is often confused with the tag O, i.e.,~Outside. This is probably due to the fact that in Bulgarian the first names sometimes have more than one meaning, which can confuse the model. The same argument holds for the tag B-LOC, which is also often confused with the tag O. Another place for improvements would be to distinguish better between B-LOC and B-ORG, as many places and organization have identical names, or at least the identical first words. The miscellaneous entities such as the names of books or movies can also have names that are identical to those of some organizations. Even more often, Miscellaneous entities could be confused with the Other category as they contain many common Bulgarian words.

\section{Conclusion and Future Work}
\label{sec:conclusion}

We explored the potential of using morphological information to a recurrent Bi-LSTM-CRF neural network architecture with the aim to improve named entity recognition for morphologically rich languages such as Bulgarian, which pose different challenges for named entity recognition compared to English. Our experiments have shown that adding morphological and part-of-speech information to the model's input yields sizable performance gains over a model that only relies on word-level and character-level embeddings as an input to the neural network.

In future work, we plan to extend the modeling of the morphological structure of the entities. Here, we only used a limited number of features, namely gender, number and definiteness, but it might be interesting to add the full linguistic knowledge encoded in the BulTreebank.
We further plan to explore features and models that can help identify loan words in Bulgarian. 

Another promising research direction is to compare the differences in the graphical representation of named entities in Bulgarian and English. For example, in English all components of a named entity are capitalized (except for the functional words). In order to have comparable data, we envision to pre-transform the Bulgarian dataset to which to apply the English capitalization rule for the phrasal named entities. 

Finally, we plan to experiment with different monolingual representations from ElMo \cite{Peters:2018}, BERT~\cite{devlin2018bert}, ROBERTa~\cite{DBLP:journals/corr/abs-1907-11692}, XLNet~\cite{DBLP:journals/corr/abs-1906-08237}, and Ernie 2.0~\cite{DBLP:journals/corr/abs-1907-12412}, pooled representations from Flair~\cite{akbik-etal-2019-pooled}, distilled representations from MT-DNN~\cite{DBLP:journals/corr/abs-1904-09482,DBLP:journals/corr/abs-1901-11504} or cross-language representations from XLM~\cite{DBLP:journals/corr/abs-1901-07291}.

\section{Acknowledgements}
This research was partially supported by {\em the Bulgarian National Interdisciplinary Research e-Infrastructure for Resources and Technologies in favor of the Bulgarian Language and Cultural Heritage, part of the EU infrastructures CLARIN and DARIAH – CLaDA-BG}, Grant number DO01-164/28.08.2018

We would like to thank the anonymous reviewers for their constructive comments, which have helped us improve the paper. 

\bibliography{bibliography}

\begin{thebibliography}{}
\expandafter\ifx\csname natexlab\endcsname\relax\def\natexlab#1{#1}\fi

\bibitem[{Akbik et~al.(2019)Akbik, Bergmann, and
  Vollgraf}]{akbik-etal-2019-pooled}
Alan Akbik, Tanja Bergmann, and Roland Vollgraf. 2019.
\newblock Pooled contextualized embeddings for named entity recognition.
\newblock In {\em Proceedings of the 2019 Conference of the North {A}merican
  Chapter of the Association for Computational Linguistics: Human Language
  Technologies\/}. Minneapolis, MN, USA, NAACL-HLT~'19, pages 724--728.

\bibitem[{Appelt et~al.(1995)Appelt, Hobbs, Bear, Israel, Kameyama, Martin,
  Myers, and Tyson}]{Appelt:1995:SIF:1072399.1072420}
Douglas~E. Appelt, Jerry~R. Hobbs, John Bear, David Israel, Megumi Kameyama,
  David Martin, Karen Myers, and Mabry Tyson. 1995.
\newblock {SRI} international {FASTUS} system: {MUC}-6 test results and
  analysis.
\newblock In {\em Proceedings of the 6th Conference on Message
  Understanding\/}. Columbia, MD, USA, MUC6~'95, pages 237--248.

\bibitem[{Bojanowski et~al.(2017)Bojanowski, Grave, Joulin, and
  Mikolov}]{BojanowskiGJM16}
Piotr Bojanowski, Edouard Grave, Armand Joulin, and Tomas Mikolov. 2017.
\newblock Enriching word vectors with subword information.
\newblock {\em Transactions of the Association for Computational Linguistics\/}
  5:135--146.

\bibitem[{Chiu and Nichols(2016)}]{Chiu2016NamedER}
Jason~P.C. Chiu and Eric Nichols. 2016.
\newblock Named entity recognition with bidirectional {LSTM}-{CNN}s.
\newblock {\em Transactions of the Association for Computational Linguistics\/}
  4:357--370.

\bibitem[{Devlin et~al.(2019)Devlin, Chang, Lee, and
  Toutanova}]{devlin2018bert}
Jacob Devlin, Ming-Wei Chang, Kenton Lee, and Kristina Toutanova. 2019.
\newblock {BERT}: Pre-training of deep bidirectional transformers for language
  understanding.
\newblock In {\em Proceedings of the 2019 Conference of the North {A}merican
  Chapter of the Association for Computational Linguistics\/}. Minneapolis, MN,
  USA, NAACL-HLT~'2019, pages 4171--4186.

\bibitem[{Doddington et~al.(2004)Doddington, Mitchell, Przybocki, Ramshaw,
  Strassel, and Weischedel}]{doddington-etal-2004-automatic}
George Doddington, Alexis Mitchell, Mark Przybocki, Lance Ramshaw, Stephanie
  Strassel, and Ralph Weischedel. 2004.
\newblock The automatic content extraction ({ACE}) program {--} tasks, data,
  and evaluation.
\newblock In {\em Proceedings of the Fourth International Conference on
  Language Resources and Evaluation\/}. Lisbon, Portugal, LREC~'04.

\bibitem[{Georgiev et~al.(2009)Georgiev, Nakov, Ganchev, Osenova, and
  Simov}]{Georgiev2009FeatureRichNE}
Georgi Georgiev, Preslav Nakov, Kuzman Ganchev, Petya Osenova, and Kiril Simov.
  2009.
\newblock Feature-rich named entity recognition for {B}ulgarian using
  conditional random fields.
\newblock In {\em Proceedings of the International Conference on Recent
  Adcances in Natural Language Processing\/}. Borovets, Bulgaria, RANLP~'09,
  pages 113--117.

\bibitem[{Gillick et~al.(2016)Gillick, Brunk, Vinyals, and
  Subramanya}]{Gillick2016MultilingualLP}
Dan Gillick, Cliff Brunk, Oriol Vinyals, and Amarnag Subramanya. 2016.
\newblock Multilingual language processing from bytes.
\newblock In {\em Proceedings of the 2016 Conference of the North American
  Chapter of the Association for Computational Linguistics: Human Language
  Technologies\/}. San Diego, CA, USA, NAACL-HLT~'16, pages 1296--1306.

\bibitem[{Grishman and Sundheim(1996)}]{Grishman:1996:MUC:992628.992709}
Ralph Grishman and Beth Sundheim. 1996.
\newblock Message understanding conference-6: A brief history.
\newblock In {\em Proceedings of the 16th Conference on Computational
  Linguistics\/}. Copenhagen, Denmark, COLING '96, pages 466--471.

\bibitem[{Hinton et~al.(2012)Hinton, Srivastava, Krizhevsky, Sutskever, and
  Salakhutdinov}]{DBLP:journals/corr/abs-1207-0580}
Geoffrey~E. Hinton, Nitish Srivastava, Alex Krizhevsky, Ilya Sutskever, and
  Ruslan Salakhutdinov. 2012.
\newblock Improving neural networks by preventing co-adaptation of feature
  detectors.
\newblock {\em CoRR\/} abs/1207.0580.

\bibitem[{Kingma and Ba(2014)}]{DBLP:journals/corr/KingmaB14}
Diederik~P. Kingma and Jimmy Ba. 2014.
\newblock Adam: {A} method for stochastic optimization.
\newblock {\em CoRR\/} abs/1412.6980.

\bibitem[{Lafferty(2001)}]{Lafferty01conditionalrandom}
John Lafferty. 2001.
\newblock Conditional {R}andom {F}ields: {P}robabilistic {M}odels for
  {S}egmenting and {L}abeling {S}equence {D}ata.
\newblock In {\em Proceedings of the 18th {I}nternational {C}onference on
  {M}achine {L}earning\/}. Morgan Kaufmann, ICML~'01, pages 282--289.

\bibitem[{Lample et~al.(2016)Lample, Ballesteros, Subramanian, Kawakami, and
  Dyer}]{N16-1030}
Guillaume Lample, Miguel Ballesteros, Sandeep Subramanian, Kazuya Kawakami, and
  Chris Dyer. 2016.
\newblock Neural architectures for named entity recognition.
\newblock In {\em Proceedings of the 2016 Conference of the North American
  Chapter of the Association for Computational Linguistics: Human Language
  Technologies\/}. San Diego, CA, USA, NAACL-HLT~'16, pages 260--270.

\bibitem[{Lample and Conneau(2019)}]{DBLP:journals/corr/abs-1901-07291}
Guillaume Lample and Alexis Conneau. 2019.
\newblock Cross-lingual language model pretraining.
\newblock {\em CoRR\/} abs/1901.07291.

\bibitem[{Le et~al.(2018)Le, Arkhipov, and
  Burtsev}]{10.1007/978-3-319-71746-3_8}
The~Anh Le, Mikhail~Y. Arkhipov, and Mikhail~S. Burtsev. 2018.
\newblock Application of a hybrid {Bi-LSTM-CRF} model to the task of {R}ussian
  named entity recognition.
\newblock In Andrey Filchenkov, Lidia Pivovarova, and Jan {\v{Z}}i{\v{z}}ka,
  editors, {\em Artificial Intelligence and Natural Language\/}. Springer
  International Publishing, pages 91--103.

\bibitem[{Ling et~al.(2015)Ling, Dyer, Black, Trancoso, Fermandez, Amir,
  Marujo, and Lu{\'\i}s}]{DBLP:conf/emnlp/LingDBTFAML15}
Wang Ling, Chris Dyer, Alan~W Black, Isabel Trancoso, Ram{\'o}n Fermandez,
  Silvio Amir, Lu{\'\i}s Marujo, and Tiago Lu{\'\i}s. 2015.
\newblock Finding function in form: Compositional character models for open
  vocabulary word representation.
\newblock In {\em Proceedings of the 2015 Conference on Empirical Methods in
  Natural Language Processing\/}. Lisbon, Portugal, EMNLP~'15, pages
  1520--1530.

\bibitem[{Liu et~al.(2019{\natexlab{a}})Liu, He, Chen, and
  Gao}]{DBLP:journals/corr/abs-1904-09482}
Xiaodong Liu, Pengcheng He, Weizhu Chen, and Jianfeng Gao. 2019{\natexlab{a}}.
\newblock Improving multi-task deep neural networks via knowledge distillation
  for natural language understanding.
\newblock {\em CoRR\/} abs/1904.09482.

\bibitem[{Liu et~al.(2019{\natexlab{b}})Liu, He, Chen, and
  Gao}]{DBLP:journals/corr/abs-1901-11504}
Xiaodong Liu, Pengcheng He, Weizhu Chen, and Jianfeng Gao. 2019{\natexlab{b}}.
\newblock Multi-task deep neural networks for natural language understanding.
\newblock {\em CoRR\/} abs/1901.11504.

\bibitem[{Liu et~al.(2019{\natexlab{c}})Liu, Ott, Goyal, Du, Joshi, Chen, Levy,
  Lewis, Zettlemoyer, and Stoyanov}]{DBLP:journals/corr/abs-1907-11692}
Yinhan Liu, Myle Ott, Naman Goyal, Jingfei Du, Mandar Joshi, Danqi Chen, Omer
  Levy, Mike Lewis, Luke Zettlemoyer, and Veselin Stoyanov. 2019{\natexlab{c}}.
\newblock {RoBERTa}: {A} robustly optimized {BERT} pretraining approach.
\newblock {\em CoRR\/} abs/1907.11692.

\bibitem[{Mikolov et~al.(2013)Mikolov, Chen, Corrado, and
  Dean}]{DBLP:journals/corr/abs-1301-3781}
Tomas Mikolov, Kai Chen, Greg Corrado, and Jeffrey Dean. 2013.
\newblock Efficient estimation of word representations in vector space.
\newblock {\em CoRR\/} abs/1301.3781.

\bibitem[{Passos et~al.(2014)Passos, Kumar, and McCallum}]{Passos2014LexiconIP}
Alexandre Passos, Vineet Kumar, and Andrew McCallum. 2014.
\newblock Lexicon infused phrase embeddings for named entity resolution.
\newblock In {\em Proceedings of the Eighteenth Conference on Computational
  Natural Language Learning\/}. Ann Arbor, MI, USA, CoNLL~'14, pages 78--86.

\bibitem[{Pennington et~al.(2014)Pennington, Socher, and
  Manning}]{pennington2014glove}
Jeffrey Pennington, Richard Socher, and Christopher Manning. 2014.
\newblock {GloVe}: Global vectors for word representation.
\newblock In {\em Proceedings of the Conference on Empirical Methods in Natural
  Language Processing\/}. Doha, Qatar, EMNLP~'14, pages 1532--1543.

\bibitem[{Peters et~al.(2018)Peters, Neumann, Iyyer, Gardner, Clark, Lee, and
  Zettlemoyer}]{Peters:2018}
Matthew Peters, Mark Neumann, Mohit Iyyer, Matt Gardner, Christopher Clark,
  Kenton Lee, and Luke Zettlemoyer. 2018.
\newblock Deep contextualized word representations.
\newblock In {\em Proceedings of the 2018 Conference of the North {A}merican
  Chapter of the Association for Computational Linguistics: Human Language
  Technologies\/}. New Orleans, LA, USA, NAACL-HLT~'18, pages 2227--2237.

\bibitem[{Piskorski et~al.(2004)Piskorski, Homola, Marciniak, Mykowiecka,
  Przepi{\'o}rkowski, and Woli{\'{n}}ski}]{10.1007/978-3-540-39985-8_24}
Jakub Piskorski, Peter Homola, Ma{\l}gorzata Marciniak, Agnieszka Mykowiecka,
  Adam Przepi{\'o}rkowski, and Marcin Woli{\'{n}}ski. 2004.
\newblock Information extraction for {P}olish using the {SProUT} platform.
\newblock In Mieczys{\l}aw~A. K{\l}opotek, S{\l}awomir~T. Wierzcho{\'{n}}, and
  Krzysztof Trojanowski, editors, {\em Intelligent Information Processing and
  Web Mining\/}. Springer Berlin Heidelberg, pages 227--236.

\bibitem[{Piskorski et~al.(2019)Piskorski, Laskova, Marci{\'n}czuk, Pivovarova,
  P{\v{r}}ib{\'a}{\v{n}}, Steinberger, and
  Yangarber}]{piskorski-etal-2019-second}
Jakub Piskorski, Laska Laskova, Micha{\l} Marci{\'n}czuk, Lidia Pivovarova,
  Pavel P{\v{r}}ib{\'a}{\v{n}}, Josef Steinberger, and Roman Yangarber. 2019.
\newblock The second cross-lingual challenge on recognition, normalization,
  classification, and linking of named entities across {S}lavic languages.
\newblock In {\em Proceedings of the 7th Workshop on Balto-Slavic Natural
  Language Processing\/}. Florence, Italy, BSNLP~'19, pages 63--74.

\bibitem[{Piskorski et~al.(2017)Piskorski, Pivovarova, {\v{S}}najder,
  Steinberger, and Yangarber}]{DBLP:conf/acl-bsnlp/PiskorskiPSSY17}
Jakub Piskorski, Lidia Pivovarova, Jan {\v{S}}najder, Josef Steinberger, and
  Roman Yangarber. 2017.
\newblock The first cross-lingual challenge on recognition, normalization, and
  matching of named entities in {S}lavic languages.
\newblock In {\em Proceedings of the 6th Workshop on {B}alto-{S}lavic Natural
  Language Processing\/}. Valencia, Spain, BSNLP~'17, pages 76--85.

\bibitem[{Popescu and Etzioni(2005)}]{Popescu:2005:EPF:1220575.1220618}
Ana-Maria Popescu and Oren Etzioni. 2005.
\newblock Extracting product features and opinions from reviews.
\newblock In {\em Proceedings of the Conference on Human Language Technology
  and Empirical Methods in Natural Language Processing\/}. Vancouver, Canada,
  EMNLP~'05, pages 339--346.

\bibitem[{Sak et~al.(2014)Sak, Senior, and
  Beaufays}]{DBLP:journals/corr/SakSB14}
Hasim Sak, Andrew~W. Senior, and Fran{\c{c}}oise Beaufays. 2014.
\newblock Long short-term memory based recurrent neural network architectures
  for large vocabulary speech recognition.
\newblock {\em CoRR\/} abs/1402.1128.

\bibitem[{Saleh et~al.(2014)Saleh, Cyphers, Glass, Joty, M\`{a}rquez,
  Moschitti, and Nakov}]{saleh-EtAl:2014:Coling}
Iman Saleh, Scott Cyphers, Jim Glass, Shafiq Joty, Llu\'{i}s M\`{a}rquez,
  Alessandro Moschitti, and Preslav Nakov. 2014.
\newblock A study of using syntactic and semantic structures for concept
  segmentation and labeling.
\newblock In {\em Proceedings of the 25th International Conference on
  Computational Linguistics\/}. Dublin, Ireland, COLING~'14, pages 193--202.

\bibitem[{Simov et~al.(2004{\natexlab{a}})Simov, Osenova, Simov, and
  Kouylekov}]{BulTreeBank}
Kiril Simov, Petya Osenova, Alexander Simov, and Milen Kouylekov.
  2004{\natexlab{a}}.
\newblock Design and implementation of the {B}ulgarian {HPSG}-based treebank.
\newblock In {\em Journal of Research on Language and Computation, Special
  Issue\/}. Kluwer Academic Publishers, pages 495--522.

\bibitem[{Simov et~al.(2004{\natexlab{b}})Simov, Osenova, and
  Slavcheva}]{BTB-TR03}
Kiril Simov, Petya Osenova, and Milena Slavcheva. 2004{\natexlab{b}}.
\newblock {BTB-TR03}: Bul{T}ree{B}ank {M}orphosyntactic {T}agset.
\newblock Bul{T}ree{B}ank {P}roject, IICT-BAS.

\bibitem[{Strakov{\'a} et~al.(2013)Strakov{\'a}, Straka, and
  Haji{\v{c}}}]{10.1007/978-3-642-40585-3_10}
Jana Strakov{\'a}, Milan Straka, and Jan Haji{\v{c}}. 2013.
\newblock A new state-of-the-art {C}zech named entity recognizer.
\newblock In Ivan Habernal and V{\'a}clav Matou{\v{s}}ek, editors, {\em Text,
  Speech, and Dialogue\/}. Springer Berlin Heidelberg, pages 68--75.

\bibitem[{Strubell et~al.(2017)Strubell, Verga, Belanger, and
  McCallum}]{Strubell:2017}
Emma Strubell, Patrick Verga, David Belanger, and Andrew McCallum. 2017.
\newblock Fast and accurate entity recognition with iterated dilated
  convolutions.
\newblock In {\em Proceedings of the Conference on Empirical Methods in Natural
  Language Processing\/}. Copenhagen, Denmark, EMNLP~'17, pages 2670--2680.

\bibitem[{Sun et~al.(2019)Sun, Wang, Li, Feng, Tian, Wu, and
  Wang}]{DBLP:journals/corr/abs-1907-12412}
Yu~Sun, Shuohuan Wang, Yukun Li, Shikun Feng, Hao Tian, Hua Wu, and Haifeng
  Wang. 2019.
\newblock {ERNIE} 2.0: {A} continual pre-training framework for language
  understanding.
\newblock {\em CoRR\/} abs/1907.12412.

\bibitem[{Sutton and McCallum(2012)}]{Sutton:2012:ICR:2688180.2688181}
Charles Sutton and Andrew McCallum. 2012.
\newblock An introduction to conditional random fields.
\newblock {\em Found. Trends Mach. Learn.\/} 4(4):267--373.

\bibitem[{Tjong Kim~Sang(2002)}]{TjongKimSang:2002:ICS:1118853.1118877}
Erik~F. Tjong Kim~Sang. 2002.
\newblock Introduction to the {CoNLL-2002} shared task: Language-independent
  named entity recognition.
\newblock In {\em Proceedings of the 6th Conference on Natural Language
  Learning\/}. Taipei, Taiwan, COLING~'02, pages 1--4.

\bibitem[{Tjong Kim~Sang and
  De~Meulder(2003)}]{TjongKimSang:2003:ICS:1119176.1119195}
Erik~F. Tjong Kim~Sang and Fien De~Meulder. 2003.
\newblock Introduction to the {CoNLL}-2003 shared task: Language-independent
  named entity recognition.
\newblock In {\em Proceedings of the Seventh Conference on Natural Language
  Learning\/}. Edmonton, Canada, CoNLL~'03, pages 142--147.

\bibitem[{Wilson et~al.(2017)Wilson, Roelofs, Stern, Srebro, and
  Recht}]{DBLP:conf/nips/WilsonRSSR17}
Ashia~C. Wilson, Rebecca Roelofs, Mitchell Stern, Nati Srebro, and Benjamin
  Recht. 2017.
\newblock The marginal value of adaptive gradient methods in machine learning.
\newblock In {\em Proceedings of the Conference on Neural Information
  Processing Systems\/}. Long Beach, CA, USA, NIPS~'17, pages 4151--4161.

\bibitem[{Yang et~al.(2019)Yang, Dai, Yang, Carbonell, Salakhutdinov, and
  Le}]{DBLP:journals/corr/abs-1906-08237}
Zhilin Yang, Zihang Dai, Yiming Yang, Jaime~G. Carbonell, Ruslan Salakhutdinov,
  and Quoc~V. Le. 2019.
\newblock {XLNet}: Generalized autoregressive pretraining for language
  understanding.
\newblock {\em CoRR\/} abs/1906.08237.

\bibitem[{Zhou and Su(2002)}]{Zhou:2002:NER:1073083.1073163}
GuoDong Zhou and Jian Su. 2002.
\newblock Named entity recognition using an {HMM}-based chunk tagger.
\newblock In {\em Proceedings of the 40th Annual Meeting on Association for
  Computational Linguistics\/}. Philadelphia, PA, USA, ACL~'02, pages 473--480.

\end{thebibliography}
\bibliographystyle{acl_natbib}

\end{document}